\documentclass[preprint,11pt]{elsarticle}
\usepackage{color,amssymb,subfigure,graphicx,amsmath,slashbox,multirow,multicol}
\usepackage{psfrag}
\usepackage{algorithm}
\usepackage{algorithmic}

\journal{arXiv}
\begin{document}
\graphicspath{{figures/}}
\newcommand{\sps}{\scriptsize}
\newcommand{\eqs}{\normalsize}
\begin{frontmatter}
\title{Classification of EEG Signal based on non-Gaussian Neutral Vector}
 \author{Zhanyu~Ma}
 \address{Pattern Recognition and Intelligent System Laboratory\\ Beijing University of Posts and Telecommunications, Beijing, China.}

\begin{abstract}
In the design of brain-computer interface systems, classification of Electroencephalogram (EEG) signals is the essential part and a challenging task. Recently, as the marginalized discrete wavelet transform (mDWT) representations can reveal features related to the transient nature of the EEG signals, the mDWT coefficients have been frequently used in EEG signal classification. In our previous work, we have proposed a super-Dirichlet distribution-based classifier, which utilized the nonnegative and sum-to-one properties of the mDWT coefficients. The proposed classifier performed better than the state-of-the-art support vector machine-based classifier. In this paper, we further study the neutrality of the mDWT coefficients. Assuming the mDWT vector coefficients to be a neutral vector, we transform them non-linearly into a set of independent scalar coefficients. Feature selection strategy is proposed on the transformed feature domain. Experimental results show that the feature selection strategy helps improving the classification accuracy.
\end{abstract}

\end{frontmatter}

%%%%%%%%%%%%%%%%%%%%%%%%%%%%%%%%%%%%%%%%%%

\section{Introduction}

Brain-computer interface (BCI) connects persons suffering from neuromuscular diseases with computers by analyzing the recorded brain signals. With a well-designed BCI system, persons with neuromuscular disease can communicate with computers enabling them to get assistances from machines. As non-invasively acquired signal, the Electroencephalogram (EEG) signal is the most studied and applied one in the design of a BCI system~\cite{Lotte2007,Chiang2012}. While a person is imagining a kind of action, the electrical activity along the scalp is recorded in the EEG signal. EEG signals show different patterns for different actions. Hence, the type of imagined action can be estimated by analyzing the EEG signals. Appropriate classification of EEG signals plays an essential role in a BCI system~\cite{Prasad2011}.

Various types of features have been extracted from EEG signals for the purpose of classification, such as the auto-aggressive (AR) parameters~\cite{Penny2000}, the multi-variate AR parameters~\cite{Chiang2012}, the Fourier transform based features~\cite{Veluvolu2012,Wang2013}, and the marginalized discrete wavelet transform (mDWT) coefficients~\cite{Subasi2007,Farina2007,Ma2012}. The DWT coefficients present the signal by projecting it onto a set of spaces. The wavelet transform applied to the EEG signal can reveal features related to the transient nature of the signal in which the time-scale regions are defined~\cite{Subasi2007}. In order to make the DWT coefficients insensitive to time alignment, the marginalized DWT (mDWT) coefficients are usually used as the feature for the task of EEG signal classification~\cite{Prasad2011,Subasi2007,Farina2007}. In this paper, we focus studying the EEG classification performance only on the mDWT features. A widely applied method, among others, is to design a classifier based on the support vector machine (SVM)~\cite{Subasi2007,Farina2007,Chang2011,08,09}. Generally speaking, the SVM-based classifier is not sensitive to the curse of dimensionality. It is also not sensitive to overtraining when choosing proper parameters~\cite{Prasad2011}. Moreover, it can easily be implemented for binary classification and extended to a multiple classes case. By involving a kernel function (\emph{e.g.}, Gaussian kernel), the performance of the SVM-based classifier could be further improved.

In EEG signal classification, the SVM-based classifier has been demonstrated as a successful tool~\cite{Subasi2010,Prasad2011}. Nevertheless, the SVM-based method does not exploit the nonnegativity and the sum-to-one nature of the mDWT coefficients~\cite{Ma2012}. In order to capture such properties, we applied the Dirichlet distribution to model the mDWT coefficients' underlying distribution. For the mDWT coefficients from more mutually independent channels, it is natural to apply the so-called super-Dirichlet distribution~\cite{Ma2011b}. In~\cite{Ma2012}, we have designed a super-Dirichlet distribution-based classifier to classify the EEG signals with mDWT representation\footnotemark\footnotetext{A super-Dirichlet variable is obtained by cascading several Dirichlet variables.}. The performance of the proposed classifier is superior to the SVM-based classifier.

It is well-known that the Dirichlet variable is a neutral vector~\cite{Connor1969,James1980}. For a vector $\mathbf{x}=[x_1,x_2,\ldots,x_{K+1}]^{\text{T}}$, an element $x_k$ is neutral if $x_1,\ldots,x_k$ is independent of $[\frac{x_{k+1}}{1 - \sum_{i=1}^k x_i},\ldots,$ $\frac{x_{K+1}}{1 - \sum_{i=1}^k x_i}]^{\text{T}}$. If all the elements in $\mathbf{x}$ are neutral, then $\mathbf{x}$ is defined as a~\emph{completely} neutral vector~\cite{Connor1969,Hankin2010}. The idea of neutrality was introduced by Connor et al.~\cite{Connor1969} to describe constrained variables with the property mentioned above. It was originally developed for biological applications. The neutral vector is highly negatively correlated. As all the elements in a neutral vector have bounded support and are nonnegative, the neutral vector cannot be described efficiently by Gaussian distribution~\cite{Ma2013}. Thus, the conventional principal component analysis (PCA) method~\cite{Bishop2006} cannot be applied for optimal decorrelation\footnotemark\footnotetext{Even though we could apply the PCA directly
to the neutral random vector variable, this linear transformation could only decorrelate the data, but can not guarantee the independence if the data is not Gaussian distributed.}. We use the parallel nonlinear transformation (PNT) to decorrelate the neutral vector in an optimal manner~\cite{Ma2013}. With such procedure, a neutral vector is decorrelated into a set of independent scalars. Moreover, if the neutral vector is treated as a vector variable and assumed to be Dirichlet distributed, the obtained scalar variables are all beta distributed~\cite{Ma2011a}. After decorrelation, we propose a feature selection strategy to keep the relevant features. Both the variance and differential entropy of the decorrelated scalar variable are used as criteria to determine which dimension should be kept.

The purpose of dimension reduction is  to remove the redundant dimensions and thus improve the corresponding performance~\cite{Bishop2006,Saeys2007,Kwak2002,He2011,Zhu2010}. We apply the proposed feature selection method in EEG signal classification tasks. The mDWT coefficients from each recording channel are assumed to be Dirichlet distributed~\cite{Ma2012,Ma2014} and decorrelated into a set of mutually independent scalars that are beta distributed. By retaining the most relevant features, we design a multi-variate beta distribution classifier for EEG signals. Experimental results demonstrate that the proposed method performs better than both the state-of-the-art SVM-based classifier~\cite{Prasad2011} and our previously proposed super-Dirichlet disribution-based classifier~\cite{Ma2012}.

The rest of this paper is organized as follows: the EEG signals are introduced in Sec.~\ref{Chap:EEG}. In Sec.~\ref{Chap:Classification via FS}, we design a classifier via feature selection. Experimental results are shown in Sec.~\ref{Chap:Exp} and some conclusions are draw in Sec.~\ref{Chap:Conclusion}

\section{Electroencephalogram Signal Analysis}
\label{Chap:EEG}

EEG signal represents the brain electrical activities over a short period of time and it is recorded from multiple electrodes placed on the scalp. Therefore, the EEG signals are obtained from multiple channels. When a classifier trained on the first day is used to classify the data from the following days, it is very difficult and challenging to achieve good performance. The EEG signal we use in this paper is obtained from the BCI competition III~\cite{BCI}. The training data and the test data were recorded from the same subject and with the same task, but on two different days with about one week in between. This way of recording data is robust to time variant.

\subsection{Data Description}

During the EEG signal recording, a subject had to perform imagined movements of either the left small finger or the tongue~\cite{BCI}. Thus we have two classes of EEG signals and the task is binary classification. The electrical brain activity was picked up during these trials using an $8\times 8$ ECoG platinum electrode grid which was placed on the contralateral (right) motor cortex. In total, $64$ channels of EEG signals were obtained. For each channel, several trials of the imaginary brain activity were recorded. In total, $278$ trials were recorded as the labeled training set and $100$ trials were recorded as the labeled test set. In both the training set and test set, the data are evenly recorded for each imaginary movement.
\subsection{Feature Extraction}

For each trial out of $278$ in the training set, $64$ channel data of length $3000$ samples were provided. Each channel data was band pass filtered in the $7-30$ Hz range\footnotemark\footnotetext{It is also suggested in other literature that the frequency characteristic can be found in even higher frequency band~\cite{Leuthardt2004}. We use the band pass, as suggested in~\cite{Prasad2011} and~\cite{Ma2012}, purely for the purpose of making the feature extraction settings consistent with previous work.} and was then processed by a multilevel one dimensional DWT. The scaling function $\Phi\left(t\right)$ and the corresponding mother wavelet function $\Psi\left(t\right)$ are presented in~\eqref{eq: DWT}, with $h\left(n\right)$ and $g\left(n\right)$ as the low-pass and high-pass filter, respectively~\cite{Farina2007}.
\begin{equation}
\label{eq: DWT}
\begin{split}
\Phi\left(t\right) &= \sqrt{2} \sum_n h\left(n\right)\Phi\left(t-n\right)\\
\Psi\left(t\right) &= \sqrt{2} \sum_n g\left(n\right)\Phi\left(t-n\right)\\
g\left(n\right) &= \left(-1\right)^{1-n}h\left(1-n\right).\\
\end{split}
\end{equation}
After the DWT, we obtained a set of coefficients $w\left(k,j\right)$, where $k=1,\ldots,K$ is the index of decomposition level, $j=0,\ldots,L/2^k-1$ is the index for the coefficient at each level, and $L$ is the length of the data from each channel. In order to make the DWT representation insensitive to time alignment, the DWT coefficients were marginalized to so-called mDWT coefficients defined as~\cite{Farina2007}\footnotemark\footnotetext{The definition in~\cite{Farina2007} was unclear about processing the low-band data obtained at the last decomposition level. We use a different expression here to make it clearer.}
\begin{equation}
\label{eq:mDWT}
\begin{split}
c_k &= \left\{
\begin{array}{cc}
\sum_{j=0}^{L/2^k-1} |w\left(k,j\right)|&k=1,\ldots,K-1\\
\sum_{j=0}^{L/2^K-1} |w_H\left(K,j\right)|& k=K\\
\sum_{j=0}^{L/2^K-1} |w_L\left(K,j\right)|& k=K+1\\
\end{array}\right .\\
&\ \ \ \ \ \ \ \ \ \ x_k= \frac{c_k}{\sum_{k=1}^{K+1} c_k},\ k=1,\ldots,K+1,
\end{split}
\end{equation}
where $w_H$ and $w_L$ denote the high-band and low-band coefficients in the last decomposition level, respectively. The normalized coefficients were cascaded into a mDWT vector as $\mathbf{x}=\left[x_1,\ldots,x_{K+1} \right]^T$. In our case, the DWT was carried out at level $K=4$ with Daubechies $2$ wavelet. Comparative work of applying different wavelets can be found in,~\emph{e.g.},~\cite{Gandhi2011}. With such settings, the total dimensionality of the mDWT vector is five. For each trial out of $278$ in the training set, we have $64$ mDWT vectors. The same procedure was also applied to the $100$ trials in the test set.

\subsection{Channel Selection}
\label{Sec:ChannelSelection}

As mentioned above, the EEG signals were recorded independently from $64$ channels, which were located on different positions over the scalp. However, it is unclear that which channels (\emph{i.e.}, recording position) are more relevant to the imaginary task than the rest~\cite{Lal2004} and the signals recorded from irrelevant channels should be noisy for the classification task~\cite{Prasad2011}. Thus the selection of the most relevant channels would improve the classification accuracy. Since it is a binary classification task in our study, we use two criteria, the Fisher ratio (FR)~\cite{Malina1981,Chae2012} and the generalization error estimation (GEE)~\cite{Lal2004}, to select the best $m$ channels, respectively.
\subsubsection{Fisher Ratio}

In binary classification, the FR presents how strong a channel correlates with labels $\left\{-1,+1\right\}$. For a channel $m$, the Fisher ratio of this channel, with equal prior probability to each class, is defined as~\cite{Malina1981}
\begin{equation}
\label{eq:FR}
\begin{split}
&\mathrm{FR}\left(m\right) = \max_{\mathbf{d}}\frac{\mathbf{d}^\text{T} \left[\boldsymbol \mu \left(m\right)_{+1} -\boldsymbol \mu \left(m\right)_{-1}\right] \left[\boldsymbol\mu\left(m\right)_{+1} -\boldsymbol \mu\left(m\right)_{-1}\right]^\text{T} \mathbf{d}}{\mathbf{d}^\text{T}\left[\boldsymbol\Sigma\left(m\right)_{+1} + \boldsymbol\Sigma\left(m\right)_{-1}\right]\mathbf{d}},
\end{split}
\end{equation}
where $\boldsymbol\mu\left(m\right)_{j}\ \textnormal{and}\ \boldsymbol\Sigma\left(m\right)_{j},\ m=1,\ldots,64,\ j\in \left\{+1,-1\right\}$ are the mean and the covariance matrix of class $j$ in channel $m$, respectively. $\mathbf{d}$ is a vector with the same size as $\boldsymbol\mu\left(m\right)_{j}$. It represents the feature space coordinate axes. The channels with larger FRs are preferable for classification. The FRs were calculated based on the training set. Table~\ref{Tab:FRCR} lists the FRs corresponds to recording channels.
\subsubsection{Generalization Error Estimation}
\label{sec:generalized error estimation}
\begin{table}[!t]
\centering
\caption{\label{Tab:FRCR}\small Fisher ratios and classification rate (in $\%$) for different channels. The best scores are in green bold font and the worse ones are in red Italic font.}
\sps
\begin{tabular}{|c||c|c|c|c|c|c|c|c|}
\hline
\ \ Channel $\sharp$\ \ & $1$ & $2$ & $3$ & $4$ & $5$ & $6$ & $7$ & $8$ \\
\hline
FR & $\ 0.03\ $ & $\ 0.04\ $ & $\ 0.09\ $ & $\ 0.04\ $ & $\ 0.02\ $ & $\ 0.07\ $ & $\ \color{red}{\emph{0.01}}\ $ & $\ 0.02\ $ \\
CR & $\ 53.24\ $ & $\ 53.96\ $ & $\ 55.04\ $ & $\ 53.24\ $ & $\ 52.52\ $ & $\ 51.08\ $ & $\ 52.16\ $ & $\ 52.16\ $ \\
\hline
\ \ Channel $\sharp$\ \ & $9$ & $10$ & $11$ & $12$ & $13$ & $14$ & $15$ & $16$\\
\hline
FR &  $\ 0.03\ $ & $\ 0.02\ $ & $\ 0.02\ $ & $\ 0.17\ $ & $\ 0.03\ $ & $\ 0.11\ $ & $\ \color{red}{\emph{0.01}}\ $ & $\ \color{red}{\emph{0.01}}\ $\\
CR &  $\ 56.83\ $ & $\ 54.68\ $ & $\ 53.60\ $ & $\ 56.47\ $ & $\ 52.88\ $ & $\ 56.83\ $ & $\ 53.96\ $ & $\ 50.36\ $\\
\hline
Channel $\sharp$& $17$ & $18$ & $19$ & $20$ & $21$ & $22$ & $23$ & $24$ \\
\hline
FR & $0.02$ & $0.13$ & $0.03$ & $0.03$ & $0.13$ & $0.18$ & $0.04$ & $0.13$ \\
CR & $51.08$ & $57.55$ & $52.52$ & $51.80$ & $59.71$ & $60.43$ & $52.88$ & $56.12$ \\
\hline
Channel $\sharp$ & $25$ & $26$ & $27$ & $28$ & $29$ & $30$ & $31$ & $32$\\
\hline
FR & $\color{red}{\emph{0.01}}$ & $0.03$ & $0.02$ & $0.05$ & $0.40$ & $0.58$ & $0.34$ & $0.08$\\
CR & $50.72$ & $56.47$ & $\color{red}{\emph{49.28}}$ & $53.96$ & $58.27$ & $\color{green}{\mathbf{70.14}}$ & $61.87$ & $55.76$\\
\hline
Channel $\sharp$& $33$ & $34$ & $35$ & $36$ & $37$ & $38$ & $39$ & $40$ \\
\hline
FR & $0.04$ & $0.03$ & $0.02$ & $0.02$ & $0.16$ & $\color{green}{\mathbf{0.88}}$ & $0.34$ & $0.17$ \\
CR & $52.52$ & $51.80$ & $53.96$ & $52.88$ & $62.95$ & $69.42$ & $58.99$ & $55.76$ \\
\hline
Channel $\sharp$& $41$ & $42$ & $43$ & $44$ & $45$ & $46$ & $47$ & $48$\\
\hline
FR &  $0.05$ & $0.04$ & $0.03$ & $0.05$ & $0.28$ & $0.25$ & $0.10$ & $0.07$\\
CR & $52.16$ & $51.08$ & $53.24$ & $52.88$ & $62.59$ & $60.43$ & $58.27$ & $57.55$\\
\hline
Channel $\sharp$& $49$ & $50$ & $51$ & $52$ & $53$ & $54$ & $55$ & $56$ \\
\hline
FR & $0.02$ & $0.07$ & $0.08$ & $0.10$ & $0.02$ & $0.10$ & $0.05$ & $0.03$ \\
CR & $52.88$ & $53.24$ & $58.27$ & $53.96$ & $51.80$ & $55.76$ & $51.44$ & $51.44$ \\
\hline
Channel $\sharp$&  $57$ & $58$ & $59$ & $60$ & $61$ & $62$ & $63$ & $64$\\
\hline
FR &  $0.03$ & $0.02$ & $0.03$ & $0.05$ & $0.06$ & $\color{red}{\emph{0.01}}$ & $0.04$ & $0.02$\\
CR &  $54.32$ & $49.64$ & $56.12$ & $56.12$ & $59.71$ & $53.24$ & $51.44$ & $51.08$\\
\hline
\end{tabular}\\
%\vspace{5mm}
\end{table}

To select channels, the performance of the channel can also be estimated by the generalization error with $N$-folds cross validation. In the BCI competition III database, the data has already been split into the training set and test set and there is no overlap between these two sets. The evaluation of the classification rate (CR) on the training set is sufficient for estimating the channel performance. For each channel, we train a SVM-based classifier with the labeled training set. With the obtained classifier, we test the performance by the labeled training set itself. The higher the CR is, the more preferable the channel is. The CRs are also listed in Table~\ref{Tab:FRCR}.

\section{EEG Classification via Feature Selection}
\label{Chap:Classification via FS}

The channel selection methods mentioned in the above section motivate us to combine different channels to obtain better classification results. As described in~\cite{Ma2012}, for each imagined trial we cascade EEG signals from the top $m$ channels to create a super-vector. The classification task is carried out based on such super-vectors.
\subsection{Super-Dirichlet Modeling}
\begin{algorithm}[!t]
   \caption{\small Parallel Nonlinear Transformation}
   \label{alg:PNT}
\footnotesize
\begin{algorithmic}
   \STATE {\bfseries Input:} Neutral vector $\mathbf{x}=[x_1,\ldots,x_K,x_{K+1}]^{\text{T}}$
   \STATE Set $\mathbf{x}_1=\mathbf{x}$, $i=2$;
     \REPEAT
    \STATE $L = \textnormal{length}(\mathbf{x}_{i-1})-1$
    \IF {$L$ is even}
    \FOR {$l=1,l\leq L/2,l++$}
     \STATE    $x_{l,i} = x_{2l-1,i-1}+x_{2l,i-1}$
     \STATE  $u_{l,i-1} = \frac{x_{2l-1,i-1}}{x_{l,i}}$
    \ENDFOR
    \STATE $\mathbf{x}_i = [x_{1,i},\ldots,x_{l,i},x_{L+1,i-1}]^{\text{T}}$
    \STATE $\mathbf{u}_{i-1} = [u_{1,i-1},\ldots,u_{l,i-1}]^{\text{T}}$
    \ELSE
    \FOR {$l=1,l<(L+1)/2,l++$}
     \STATE $x_{l,i} = x_{2l-1,i-1}+x_{2l,i-1}$
     \STATE $u_{l,i-1} = \frac{x_{2l-1,i-1}}{x_{l,i}}$
    \ENDFOR
    \STATE $\mathbf{x}_i = [x_{1,i},\ldots,x_{l,i}]^{\text{T}}$
    \STATE $\mathbf{u}_{i-1} = [u_{1,i-1},\ldots,u_{l,i-1}]^{\text{T}}$
    \ENDIF
    \STATE $i=i+1$
   \UNTIL{$\textnormal{length}(\mathbf{x}_{i})==2$}
   \STATE Set $\mathbf{u}_i={x}_{1,i}$
   \STATE {\bfseries Output:} Transformed vector $\mathbf{u} = [\mathbf{u}_1^{\text{T}},\ldots,\mathbf{u}_{i}^{\text{T}}]^{\text{T}}$, which is of size $K$.
\end{algorithmic}
\end{algorithm}

According to~\eqref{eq:mDWT}, the mDWT vector extracted from each channel contains elements which are nonnegative and whose sum is one. Hence, it is natural to model the underlying distribution of the mDWT vector by Dirichlet distribution. For more than one channels, we apply the super-Dirichlet distribution~\cite{Ma2011b} to describe the super-vector's distribution. For a super-vector from the top $m$ channels $\mathbf{x}_{\text{sup}}=[\mathbf{x}_1^\text{T},\mathbf{x}_2^\text{T},\ldots,\mathbf{x}_m^\text{T}]^\text{T}$ ($\mathbf{x}_t=[x_1,x_2,\ldots,x_{K+1}]^\text{T}$), the probability density function (PDF) of the super-Dirichlet distribution is defined as
\begin{equation}
\begin{split}
&\mathbf{sDir}(\mathbf{x}_{\text{sup}}; \boldsymbol{\alpha}) =\prod_{t=1}^m \mathbf{Dir}(\mathbf{x}_t;\boldsymbol{\alpha}_t)=\prod_{t=1}^m \frac{\Gamma\left( \sum_{k=1}^{K_t+1} \alpha_{t,k}\right)}{\prod_{k=1}^{K_t+1}\Gamma\left( \alpha_{t,k} \right)}\prod_{k=1}^{K_t+1} x_{t,k}^{\alpha_{t,k}\ -1},
\end{split}
\end{equation}
where $\Gamma (\cdot)$ is the gamma function, $m$ is the number of subvectors (\emph{i.e.}, the number of selected channels) in the super-vector, and $K_t$ is the degrees of
freedom of the $t$th subvector (in our case, $K_1 = \cdots = K_m = 4$). $\alpha_{t,k}$ is the parameter corresponds to $x_{t,k}$, where $x_{t,k}$ denotes the $k$th element in the $t$th subvector $\mathbf{x}_t,\ t=1,\ldots,m$. The PDF of the super-Dirichlet distribution is actually a multiplication of several PDFs of the Dirichlet distribution. The parameter estimation methods for the super-Dirichlet distribution can be found in~\cite{Ma2012}.

\subsection{Non-linear Decorrelation of Neutral Vector}
\subsubsection{Neutral Vector}

Assuming we have a random vector variable $\mathbf{x}=[x_1,x_2,$ $\ldots,x_K,x_{K+1}]^{\text{T}}$, where $x_k\geq 0$ and $\sum_{k=1}^{K+1}x_k=1$.
An element $x_k$ is neutral if $x_1,\ldots,x_k$ is independent of $[\frac{x_{k+1}}{1 - \sum_{i=1}^k x_i},\ldots,$ $\frac{x_{K+1}}{1 - \sum_{i=1}^k x_i}]^{\text{T}}$. If all the elements in $\mathbf{x}$ are neutral, then $\mathbf{x}$ is defined as a~\emph{completely} neutral vector~\cite{Connor1969,Hankin2010}. A neutral vector with $K+1$ elements has $K$ degrees of freedom. According to the above definition, the neutral vector conveys a particular type of independence among its elements, even though the element variables themselves are mutually negatively correlated.

\subsubsection{Decorrelation via Parallel Non-linear Transformation }

In most signal processing applications, the transformations we use are linear or non-linear according to some nonlinear kernel functions. Even though we could apply PCA directly to the neutral random vector variable, this linear transformation could only decorrelate the data, but cannot guarantee the independence if the data is not Gaussian. Furthermore, the PCA does not exploit the neutrality~\cite{Ma2011}. Therefore, PCA is not optimal for decorrelating neutral vector. By considering the neutrality, we apply nonlinear invertible transformation in this paper, which decorrelates the vector variable into a set of mutually independent variables. In contrast to PCA, the transformations do not require any statistical information (\emph{e.g.}, the covariance matrix) of the observed vector set. Thus, it avoids the eigenvalue analysis for PCA and, therefore, the computational cost is saved.

As each element in $\mathbf{x}$ is neutral, with the neutrality of $x_1$, we know that $x_1$ is independent of the remaining normalized elements. The remaining normalized elements then build a new neutral vector. Based on this fact, the parallel non-linear transformation (PNT) scheme described in Algorithm~\ref{alg:PNT} can be applied to non-linearly decorrelate $\mathbf{x}$ to a vector $\mathbf{u}$ with $K$ mutually independent variables. Discussion of the independence is presented in~\cite{Ma2013}. The nonlinear transformation scheme proposed above is invertible by iterative multiplications. It shows the PNT procedure for $5$ dimensional neutral vector.

\subsubsection{Distribution of the Decorrelated Elements}
\begin{algorithm}[!t]
   \caption{\small Calculation of Parameters in Beta Distributions}
   \label{alg:BetaParameter}
\footnotesize
\begin{algorithmic}
   \STATE {\bfseries Input:} Original Dirichlet parameters $\boldsymbol{\alpha}=[\alpha_1,\ldots,\alpha_K,\alpha_{K+1}]^{\text{T}}$
   \STATE Set $\boldsymbol{\alpha}_1=\boldsymbol{\alpha}$, $i=2$;
     \REPEAT
    \STATE $L = \textnormal{length}(\boldsymbol{\alpha}_{i-1})-1$
    \IF {$L$ is even}
    \FOR {$l=1,l\leq L/2,l++$}
     \STATE    $\alpha_{l,i} = \alpha_{2l-1,i-1}+\alpha_{2l,i-1}$
     \STATE  $a_{l,i-1} = \alpha_{2l-1,i-1}$, $b_{l,i-1} = \alpha_{2l,i-1}$
    \ENDFOR
    \STATE $\boldsymbol{\alpha}_i = [\alpha_{1,i},\ldots,\alpha_{l,i},\alpha_{L+1,i-1}]^{\text{T}}$
    \STATE $\mathbf{a}_{i-1} = [a_{1,i-1},\ldots,a_{l,i-1}]^{\text{T}}$, $\mathbf{b}_{i-1} = [b_{1,i-1},\ldots,b_{l,i-1}]^{\text{T}}$
    \ELSE
    \FOR {$l=1,l<(L+1)/2,l++$}
     \STATE $\alpha_{l,i} = \alpha_{2l-1,i-1}+\alpha_{2l,i-1}$
     \STATE $a_{l,i-1} = \alpha_{2l-1,i-1}$, $b_{l,i-1} = \alpha_{2l,i-1}$
    \ENDFOR
    \STATE $\boldsymbol{\alpha}_i = [\alpha_{1,i},\ldots,\alpha_{l,i}]^{\text{T}}$
    \STATE $\mathbf{a}_{i-1} = [a_{1,i-1},\ldots,a_{l,i-1}]^{\text{T}}$, $\mathbf{b}_{i-1} = [b_{1,i-1},\ldots,b_{l,i-1}]^{\text{T}}$
    \ENDIF
        \STATE $i=i+1$
   \UNTIL{$\textnormal{length}(\boldsymbol{\alpha}_{i})==2$}
   \STATE Set $\mathbf{a}_i={\alpha}_{1,i}$, $\mathbf{b}_i={\alpha}_{2,i}$
   \STATE {\bfseries Output:} Parameters for the transformed variable: $\mathbf{a} = [\mathbf{a}_1^{\text{T}},\ldots,\mathbf{a}_{i}^{\text{T}}]^{\text{T}}$ and $\mathbf{b} = [\mathbf{b}_1^{\text{T}},\ldots,\mathbf{b}_{i}^{\text{T}}]^{\text{T}}$, which are all of size $K$.
\end{algorithmic}
\end{algorithm}

The Dirichlet variable is a~\emph{completely} neutral vector~\cite{Frigyik2010}. Assuming $\mathbf{x}=[x_1,x_2,\ldots,x_K,x_{K+1}]^{\text{T}}$ is a Dirichlet variable whose PDF is $\mathbf{x}\sim \mathbf{Dir}(\mathbf{x};\boldsymbol{\alpha})$, we apply the above proposed PNT algorithm to decorrelate $\mathbf{x}$ to obtain $\mathbf{u}$. Moreover, all the elements in $\mathbf{u}$ are not only~\emph{decorrelated} but also mutually~\emph{independent}. The parameters in the Dirichlet PDF are $\boldsymbol{\alpha}=[\alpha_1,\alpha_2,\ldots,\alpha_K,\alpha_{K+1}]^{\text{T}}$. With the permutable property, aggregation property and the neutral property~\cite{Ma2013}, each element in obtained vector $\mathbf{u}$ is beta distributed. The algorithm of calculating the parameters for the resulted beta distributions are described in Algorithm~\ref{alg:BetaParameter}. For the example, we have
\begin{equation}
\begin{split}
u_1\sim \mathbf{Beta}(u_1;\alpha_1,\alpha_2);&\ \ \ \ \ u_2\sim \mathbf{Beta}(u_2;\alpha_3,\alpha_4);\\
u_3\sim \mathbf{Beta}(u_3;\alpha_1+\alpha_2,\alpha_3+\alpha_4);&\ \ \ \ \ u_4\sim \mathbf{Beta}(u_4;\sum_{i=1}^4\alpha_i,\alpha_5),
\end{split}
\end{equation}
where
\begin{equation}
\mathbf{Beta}(x;a,b) = \frac{\Gamma(a+b)}{\Gamma(a)\Gamma(b)}x^{a-1}(1-x)^{b-1}.
\end{equation}
To illustrate the decorrelation effect of the PNT schemes on the Dirichlet variable, we generated $100,000$ vectors from a Dirichlet distribution with $\boldsymbol \alpha = [2,5,6,3,7]^{\text{T}}$. xThe sample correlation coefficient $R_{x_{i,j}}$ for the original element pair $(x_i,x_j)$ was also evaluated. Table shows the sample correlation coefficients before and after transformation with PNT. The coefficients are very small after transformation, hence the correlation between each element pair vanished.
%
%\begin{figure}[!t]
%          \centering
%\begin{tabular}{@{}c c@{}}
%\subfigure[\label{ToyData}\small The joint distribution of the $1$st and $3$rd dimensions ($x_1$ and $x_3$) of synthetic data.]{
%          \includegraphics[width=.35\textwidth]{ToyData.eps}} &          \subfigure[\label{ToyDataTransformed}\small The joint distribution of the transformed coefficients ($u_1$ and $u_2$) by PNT.]{
%          \includegraphics[width=.38\textwidth]{ToyDataTransformed.eps}}\\
%
%\subfigure[\label{Tab:Original}\small The sample correlation coefficients of the original sample vectors.]{
%          \includegraphics[width=.35\textwidth]{ToyDataStat.eps}} &          \subfigure[\label{Tab:PNT}\small The sample correlation coefficients of the transformed coefficients.]{
%          \includegraphics[width=.4\textwidth]{ToyDataTransformedStat.eps}}
%\end{tabular}
%\caption{\label{DirichletExample}\small An example of decorrelating samples from a five-dimensional Dirichlet random variable with
%PNT. The data set was generated from $\mathbf{Dir}(\mathbf{x};\boldsymbol \alpha)$ with $\boldsymbol \alpha = [2,5,6,3,7]^{\text{T}}$. It is clearly shown that the correlation is removed.}
%\end{figure}

\subsection{Selection of Relevant Features}
\label{Chap:FeatureSelection}

Feature selection is an important problem in EEG signal classification~\cite{Peng2005,Prasad2011,Lawhern2013}. In section~\ref{Sec:ChannelSelection}, the FR and GEE were applied to select the most relevant channels. However, within each channel, it is unknown which dimensions are more relevant to the class labels than others. Another difficulty for feature selection within each channel is that the feature in different dimensions are highly negatively correlated. The above introduced decorrelation strategy can transform the negatively correlated Dirichlet vector variable into a set of mutually independent scalar variables. Thus, we can directly select the features without considering the correlations among them.

Typically, two criteria can be used for feature selection, the variance of the data~\cite{Bishop2006,He2011} and the differential entropy of the data~\cite{Kwak2002,Zhu2010}. The variance reflects how far a set of data are spread out. The differential entropy is a measure of average uncertainty of a random variable under continuous probability distributions. In general, the dimension with larger variance/differential entropy is preferred in classification, as they can better describe the divergence among the data. With the assumption that the source data is Dirichlet distributed, the transformed vector contains a set of scalar variables which are beta distributed. For beta distribution $\mathbf{Beta}(x;a,b)$, the variance of $x$ is computed as
\begin{equation}
\text{var}(x) = \mathbf{E}\left\{\left[x-\mathbf{E}(x)\right]^2\right\}=\frac{ab}{(a+b)^2(a+b+1)},
\end{equation}
and the differential entropy of $x$ is calculated as
\begin{equation}
\text{H}(x) = -\mathbf{E}\left[\ln \mathbf{Beta}(x;a,b)\right]=\ln\frac{\Gamma(a)\Gamma(b)}{\Gamma(a+b)}-(a-1)\psi(a) - (b-1)\psi(b) + (a+b-2)\psi(a+b),
\end{equation}
where $\psi(x)$ is the digamma function defined as $\psi(x)=\frac{\partial\ln \Gamma(x)}{\partial x}$.

In the following paragraph, we use both of the above mentioned criteria to select $R$ dimensions that correlate with the $R$ largest variances or differential entropies.

\subsection{Multi-variate Beta Distribution-based MAP Classifier}
\label{Chap:mvBetaClassifier}
According to the above procedure, a set of selected dimensions are obtained. As the data in each dimension is assumed to be beta distributed and the dimensions are mutually independent, we can model the underlying distribution of the selected $R$-dimensional vector variable $\mathbf{\tilde{u}}=[u_1,u_2,\ldots,u_R]^{\text{T}}$, which are selected from one recording channel, by a multi-variate beta distribution (mvBeta) as
\begin{equation}
f(\mathbf{\tilde{u}}) = \prod_{r=1}^ R \mathbf{Beta}(\tilde{u}_r;a_r,b_r).
\end{equation}

Similarly, for the recordings from top $m$ channels, there are $R\times m$ dimension selected in total. Therefore, these dimensions are modeled as
\begin{equation}
f(\mathbf{\tilde{u}}_{\text{sup}}) = \prod_{i=1}^m\prod_{r=1}^ R \mathbf{Beta}(\tilde{u}_{ir};a_{ir},b_{ir}),
\end{equation}
where $\mathbf{\tilde{u}}_{\text{sup}} = [\mathbf{\tilde{u}}_1^{\text{T}},\mathbf{\tilde{u}}_2^{\text{T}},\ldots,\mathbf{\tilde{u}}_m^{\text{T}}]^{\text{T}}$.

The BCI competition III data contains two classes, with label index $C\in\{+1,-1\}$. Since the parameters in the beta distributions are known according to Algorithm~\ref{alg:BetaParameter}, a class dependent mvBeta distribution can be obtained for each class. In the test procedure, we create a maximum a posterior (MAP) classifier with the above obtained models. In each recording channel, for the vector $\mathbf{x}^t$ from a test trial, we firstly transform it into $\mathbf{u}^t$ with Algorithm~\ref{alg:PNT}, and then select the $R$ dimensions via the dimension's variance/entropy. Finally, a decision based on the selected features for $m$ recording channels is made as
\begin{equation}
\label{eq:MAP Classifier}
\left\{\begin{array}{cc}
\text{if}\ f\left(C=+1|\mathbf{\tilde{u}}_{\text{sup}}^t\right) \geq f\left(C=-1|\mathbf{\tilde{u}}_{\text{sup}}^t\right) & \mathbf{\tilde{u}}_{\text{sup}}^t\in +1\\
\text{else} &  \mathbf{\tilde{u}}_{\text{sup}}^t\in -1
\end{array}\right .,
\end{equation}
where $f(C|\mathbf{\tilde{u}}_{\text{sup}}^t)\propto {f\left(\mathbf{\tilde{u}}_{\text{sup}}^t|C\right)p\left(C\right)}$.
\section{Experimental Results and Discussions}
\label{Chap:Exp}

We evaluated the performance of the proposed feature selection strategy with the mvBeta distribution-based classifier on the BCI competition III database and compared it with the SVM-based classifier, the recently proposed super-Dirichlet mixture model (sDMM)-based method, and the PCA-based classifier. The DWT is calculated using Matlab~\emph{wavedec} function with declevel equal to $4$, followed by marginalization described in~\eqref{eq:mDWT}. According to Tab.~\ref{Tab:FRCR}, the best $m$ channels were selected based on FRs or CRs, in terms of their ranks.

Classifier setting and implementations:
\begin{itemize}
\item The mvBeta-based classifier was implemented according to the description in section~\ref{Chap:mvBetaClassifier}. Feature selection was carried out within each channel.
\item The LIBSVM~\cite{Chang2011} was used to implement the SVM-based classifier, which had Gaussian kernel function with $\gamma=4$ and the soft margin parameter $C=1$. \emph{No} feature selection was applied for SVM-based classifier.
\item The sDMM-based classifier was implemented based on the method described in~\cite{Ma2012}. There was \emph{no} decorrelation strategy for sDMM or \emph{no} feature selection either.
\item The PCA-based classifier was implemented with the standard PCA method. Within each channel, PCA was applied to decorrelate the data and features were selected according to their variances. The Gaussian mixture model was applied to model the distribution of the selected features.
\end{itemize}
All the above mentioned classifiers were trained and evaluated based on mDWT coefficients collected from the best $m$ channels.

\subsection{Classification Accuracy without Feature Selection}

%\begin{figure}[!t]
%
%\vspace{0mm}
%          \centering
%          \includegraphics[width=.85\textwidth]{FR-all.eps}
%          \vspace{-0mm}
%          \caption{\label{Fig:FR-all}\small Classification rates of mvBeta based classifier, with channel selection based on FRs.}
%          \vspace{-0mm}
%\end{figure}

In order to demonstrate the non-linear decorrelation strategy, we evaluated the mvBeta distribution-based classifier~\emph{without} feature selection, which means that we set $R=K=4$. In such case, the proposed classifier should perform the same as the one used in~\cite{Ma2012}, as no information is added or lost during the non-linear transformation. As expected, experimental results show identical performance as that reported in~\cite{Ma2012}, where the sDMM-based classifier was employed. The highest classification accuracy is $75\%$ for both cases.

\subsection{Classification Accuracy with Feature Selection}

\begin{figure}[!t]
          \centering
\begin{tabular}{@{}c c@{}}
\subfigure[\label{FR-L4}\scriptsize Channel selection with Fisher ratio and $R=4$.]{
          \includegraphics[width=.45\textwidth]{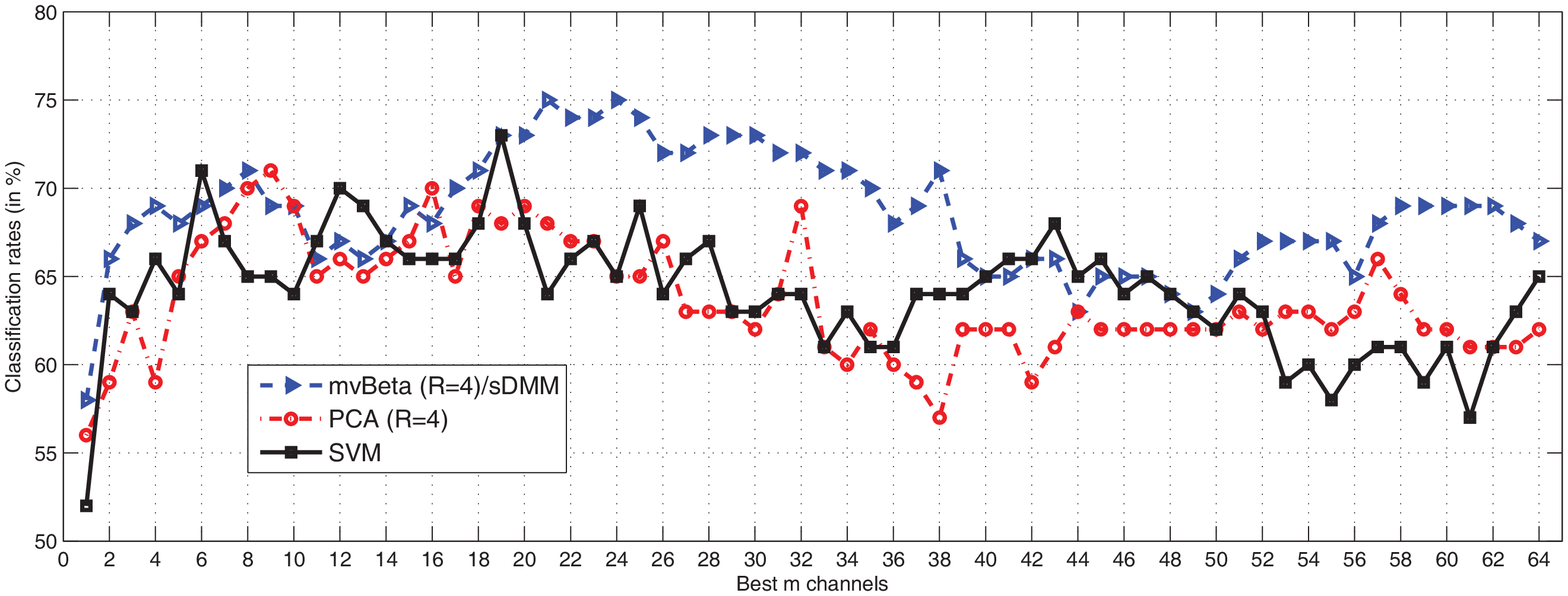}} &          \subfigure[\label{CR-L4}\scriptsize Channel selection with classification rates and $R=4$.]{
          \includegraphics[width=.45\textwidth]{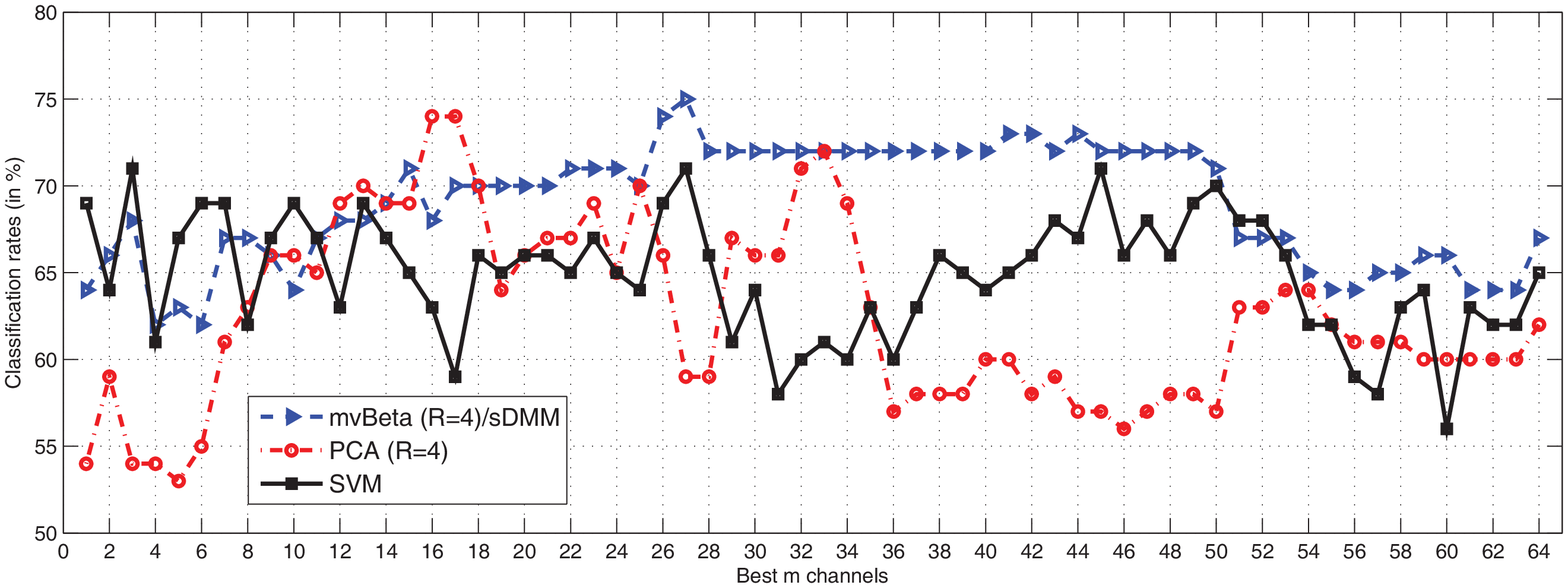}}\\
          \subfigure[\label{FR-L3}\scriptsize Channel selection with Fisher ratio and $R=3$.]{
          \includegraphics[width=.45\textwidth]{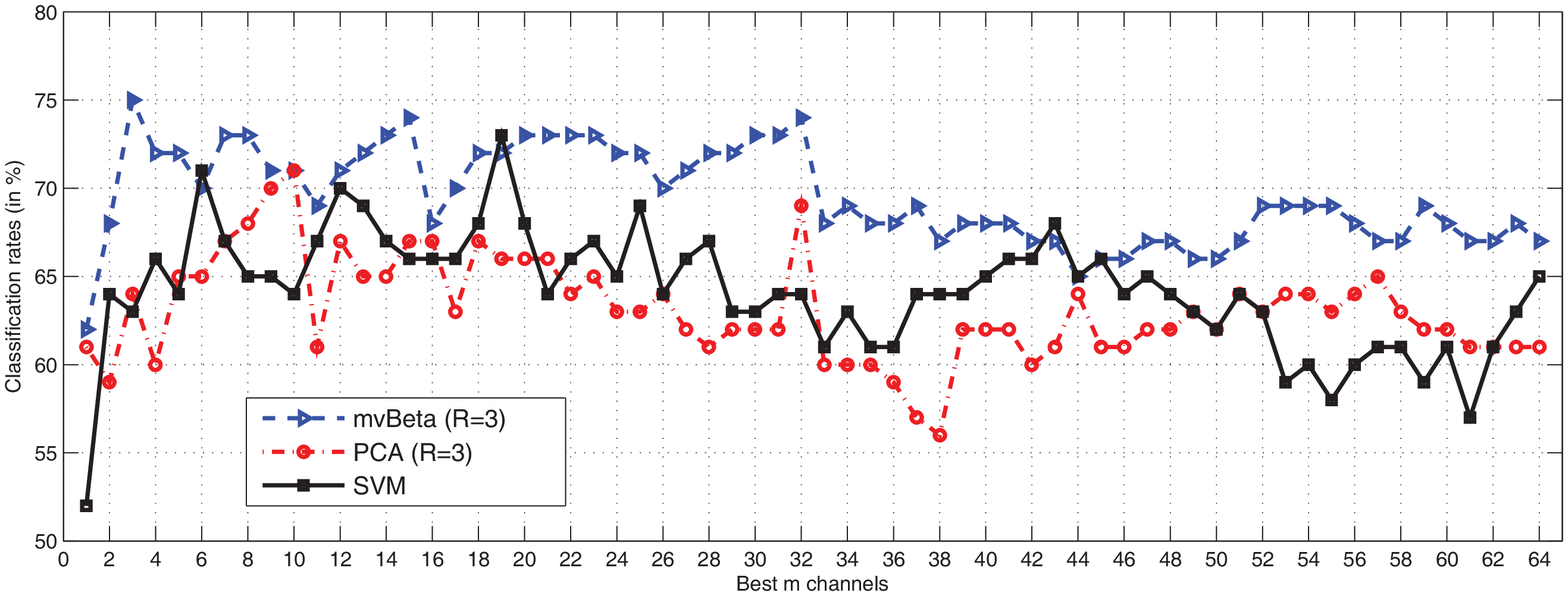}} &          \subfigure[\label{CR-L3}\scriptsize Channel selection with classification rates and $R=3$.]{
          \includegraphics[width=.45\textwidth]{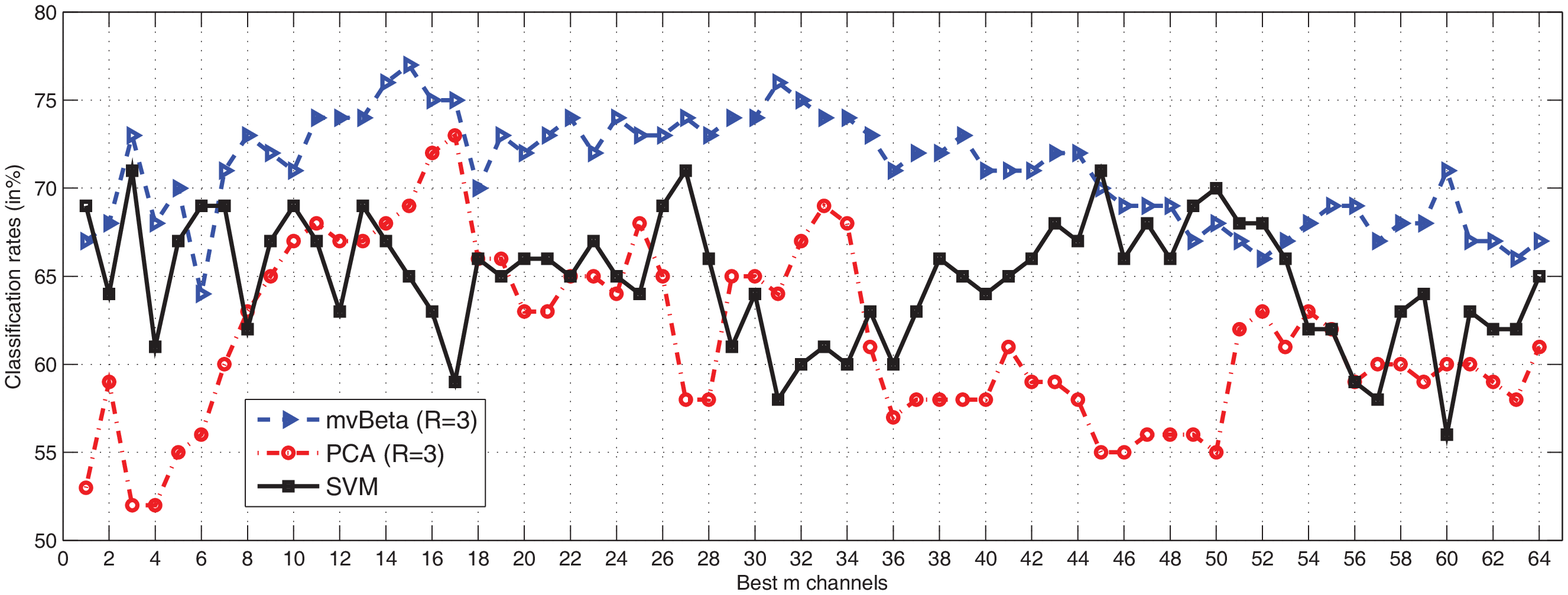}}\\
          \subfigure[\label{FR-L2}\scriptsize Channel selection with Fisher ratio and $R=2$.]{
          \includegraphics[width=.45\textwidth]{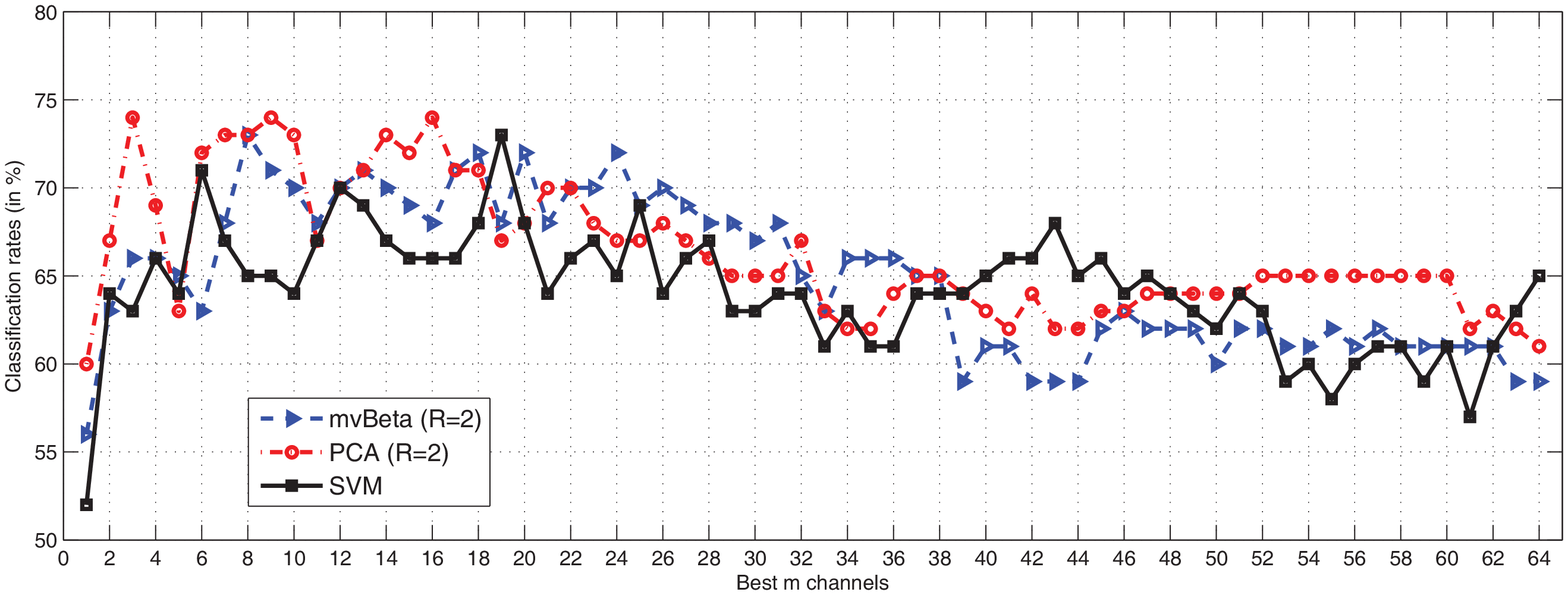}} &          \subfigure[\label{CR-L2}\scriptsize Channel selection with classification rates and $R=2$.]{
          \includegraphics[width=.45\textwidth]{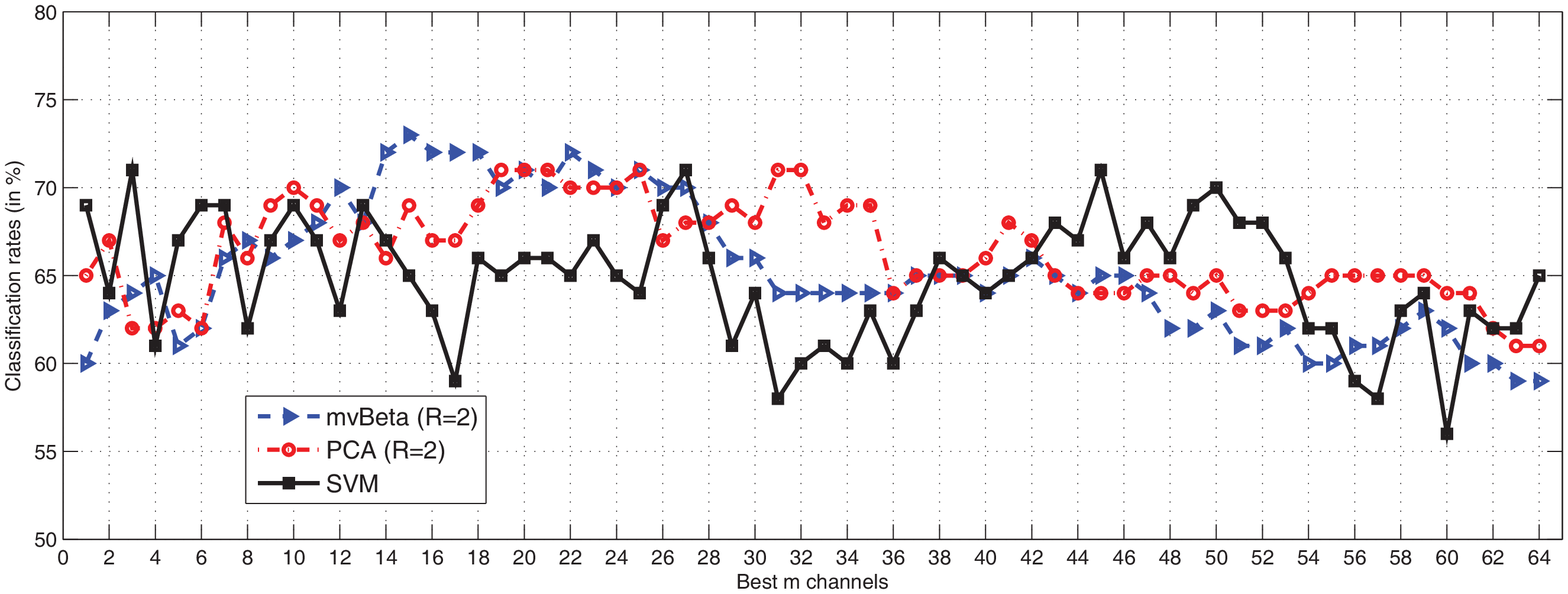}}\\
\end{tabular}
\caption{\label{Fig:Classification Rates}\small Classification rates comparisons of mvBeta-based classifier, PCA-based classifier, and SVM-based classifier, with different channel selection strategies and number of selected channels.}
\end{figure}

The total dimension of the mDWT is $5$ for each recording channel, which has the degrees of freedom equal to $4$. Hence, after decorrelation (both with PNT and PCA), the obtained vector are $4$ dimensional ($K=4$). In order to evaluate the mvBeta distribution-based classifier with the proposed feature selection strategy in Sec.~\ref{Chap:FeatureSelection}, we set $R=3$ and $R=2$, respectively\footnotemark\footnotetext{We have tried both the variance and the differential entropy criteria. For the BCI competition III data set that used in this paper, these two criteria yield exactly the same order of features.}. We also took similar feature selection choices for the PCA-based classifier. The classification accuracies are illustrated in Fig.~\ref{Fig:Classification Rates}.

It can be observed that for the FR case (Fig.~\ref{FR-L4},~\ref{FR-L3}, and~\ref{FR-L2}), when setting $R=3$, the best performance $75\%$ appears at $m=3$ for mvBeta distribution-based classifier. This classification rate is the same as that obtained by the sDMM/mvBeta distribution (without feature selection)-based classifiers, the only difference is the best performance occurs at $m=21,24$ in the latter classifiers. For the PCA-based classifier, the best performance, which is $74\%$, appears at $m=3$ with $R=2$. When investigating the CR case (Fig.~\ref{CR-L4},~\ref{CR-L3}, and~\ref{CR-L2}), it can be observed that the mvBeta distribution-based classifier performs better than the sDMM/mvBeta distribution (without feature selection)-based classifiers. The classification rate reaches $77\%$ at $m=15$ and $76\%$ at $m=31$. Meanwhile, $75\%$ has been reached at several $m$s. This fact supports our motivation that removing redundant features can improve the classification performance. The choice of $R=2$ does not work well, which is because we have reduced too much dimensions and key information are lost. The best performance of the PCA-based classifier is again $74\%$, which happens at $m=16$. In this case, feature selection does not help in improving the classification accuracy.

%\begin{figure}[!t]
%
%\vspace{0mm}
%          \centering
%          \includegraphics[width=.85\textwidth]{CR-all.eps}
%          \vspace{-0mm}
%          \caption{\label{Fig:CR-all}\small Classification rates of mvBeta based classifier, with channel selection based on CRs.}
%          \vspace{-0mm}
%\end{figure}
\begin{table}[!t]
\centering
 \caption{\label{Tab:ComparisonSummary}Summary of classification rates ($R=4$ is the case without feature selection).}
\sps
\begin{tabular}{|@{}c@{}|@{}c@{}|@{}c@{}|@{}c@{}|@{}c@{}|}
\hline
\multirow{1}{*}{Channel Selection} & \multirow{1}{*}{Classifier} & \multirow{1}{*}{Best performance} & \multirow{1}{*}{Mean Acc.}& \multirow{1}{*}{Std. Dev.}\\
\hline
\hline
\multirow{7}{*}{Fisher ratio}& mvBeta $(R=4)$/sDMM  & $\mathbf{75}\%$ ($m=21,24$)& ${68.59}\%$ & $0.0336$\\
                             & mvBeta $(R=3)$ & $\mathbf{75}\%$ ($m=3$)& $\mathbf{69.53}\%$ &$\mathbf{0.0273}$ \\
                             & mvBeta $(R=2)$ & $73\%$ ($m=8$)& $64.97\%$& $0.0431$\\
                             \cline{2-5}
                             & PCA $(R=4)$  & ${71}\%$ ($m=9$)& ${63.67}\%$ & $0.0330$\\
                             & PCA $(R=3)$ & ${71}\%$ ($m=10$)& ${63.11}\%$ &${0.0289}$ \\
                             & PCA $(R=2)$ & $74\%$ ($m=3$)& $66.31\%$& $0.0373$\\
                             \cline{2-5}
                             & SVM   & $73\%$ ($m=19$)& $64.17\%$& $0.0342$\\
                             \hline
\multirow{7}{*}  {\ \ Classification rate\ \ }&\ \ \ mvBeta $(R=4)$/sDMM\ \ \  & $75\%$ ($m=27$)&${68.98}\%$& $0.0341$\\
                             & mvBeta $(R=3)$ & $\mathbf{77}\%$ ($m=15$)& $\mathbf{71.05}\%$& $\mathbf{0.0301}$\\
                             & mvBeta $(R=2)$ & $73\%$ ($m=15$)& $65.28\%$& $0.0389$\\
                             \cline{2-5}
                             &\ \ \ PCA $(R=4)$\ \ \  & $74\%$ ($m=16$)&${62.50}\%$& $0.0526$\\
                             & PCA $(R=3)$ & ${73}\%$ ($m=17$)& ${61.44}\%$& ${0.0487}$\\
                             & PCA $(R=2)$ & $71\%$ ($m=19$)& $66.31\%$& $0.0282$\\
                             \cline{2-5}
                             & SVM  &\ \  $71\%$ ($m=3,27,45$)\ \ \ &\ \ \ $64.84\%\ \ \ $& $0.0347$\\
             \hline
\end{tabular}\\
\end{table}
\subsection{Discussion}

In general, the non-linear decorrelation strategy for the neutral vector works well in EEG signal classification, no matter with or without feature selection. This verifies the effectiveness of the non-linear decorrelation strategy.

When comparing with the SVM-based classifier~\cite{Prasad2011}, the recently proposed sDMM-based classifier~\cite{Ma2012} and the PCA-based classifier, the feature selection strategy proposed in this paper indeed improves the classification results. A summary of comparisons is listed in Tab.~\ref{Tab:ComparisonSummary}.

For the FR case, the mvBeta distribution-based classifier (with $R=3$) and the sDMM-based classifier have the same highest accuracies. However, the latter one needs to involve more channels ($m=21$  or $m=24$) while the former one obtains the same classification rate at $m=3$. This indicates that the latter method has higher complexity. Comparing with the best PCA-based classifier ($R=2$ and $m=3$), the mvBeta classification-based classifier improves the classification rate by $1\%$. The mean accuracy is improved as well. For the CR case, the mvBeta distribution-based classifier (with $R=3$) outperforms the sDMM-based classifier by $2\%$ and outperforms the PCA-based classifier ($R=4$ and $m=16$) by $3\%$. Similar to the FR case, the mvBeta distribution-based classifier requires less channels. Moreover, when comparing the mean classification rate and the standard deviation, the mvBeta distribution-based classifier (with $R=3$) is more reliable and stable than all the other methods.

To further test the statistical meaning of the classification accuracies, we also applied the Student's t-test to analyze the results. The $p$-values of the null hypothesis that the two compared methods perform similar are listed in Tab.~\ref{Tab:pValueComparisonSummary}. All the $p$-values are further smaller than $0.01$ and, therefore, the null hypothesis are rejected. This means that the proposed mvBeta distribution-based method indeed improves the classification accuracy.
\begin{table}[!t]
\centering
 \caption{\label{Tab:pValueComparisonSummary} $p$-values of the Student's t-test for the ``null hypothesis'' that the classification performance of two methods are similar. The best performance of each method is selected for comparisons.}
\scriptsize
\begin{tabular}{|c|c|c|}%{|@{}c@{}|@{}c@{}|@{}c@{}|@{}c@{}|@{}c@{}|}
\hline
   & \multicolumn{2}{c|}{Fisher ratio} \\
\hline
Null hypothesis   & mvBeta $(R=3)$ \& SVM & mvBeta$(R=3)$ \& PCA ($R=2$)\\
   \hline
   $p$-value & $4.81\times 10^{-17}$ & $1.68\times 10^{-7}$ \\
   \hline
   \hline
  & \multicolumn{2}{c|}{Classification rate}\\
\hline
Null hypothesis   & mvBeta $(R=3)$ \& SVM & mvBeta$(R=3)$ \& PCA ($R=4$)\\
   \hline
   $p$-value & $1.54\times 10^{-19}$ & $1.49\times 10^{-19}$\\
   \hline
\end{tabular}\\
\end{table}
%\begin{table}[!t]
%\centering
% \caption{\label{Tab:pValueComparisonSummary} $p$-values of the Student's t-test for the ``null hypothesis'' that the classification performance of two methods are similar. The best performance of each method is selected for comparisons.}
%\scriptsize
%\begin{tabular}{|c|c|c|c|c|}%{|@{}c@{}|@{}c@{}|@{}c@{}|@{}c@{}|@{}c@{}|}
%\hline
%Method   & \multicolumn{2}{c|}{Fisher ratio} & \multicolumn{2}{c|}{Classification rate}\\
%\hline
%Null hypothesis   & mvBeta $(R=3)$ \& SVM & mvBeta$(R=3)$ \& PCA ($R=2$)& mvBeta $(R=3)$ \& SVM & mvBeta$(R=3)$ \& PCA ($R=4$)\\
%   \hline
%   $p$-value & $4.81\times 10^{-17}$ & $1.68\times 10^{-7}$ & $1.54\times 10^{-19}$ & $1.49\times 10^{-19}$\\
%   \hline
%\end{tabular}\\
%\end{table}

\section{Conclusions and future work}
\label{Chap:Conclusion}

In order to optimally remove the correlation among the feature dimensions and thus improve classification accuracy, a parallel non-linear transformation strategy was applied to decorrelate the negatively correlated neutral vector. Specially, when the neutral vector is Dirichlet distributed, the obtained decorrelated scalar variables are mutually independent and each of them is beta distributed. After decorrelation, we applied the variance and the differential entropy as criteria in feature selection. The proposed feature selection strategy with non-linear transformation has been employed in EEG signal classification. Experimental results demonstrate that classifier based on the selected features performs better and is more stable than the SVM-based classifier, the recently proposed sDMM-based classifier, and the PCA-based classifier.

There are many possible ways to improve the classification accuracy in the future work. In current work, the feature selection is conducted for each channel independently. If we apply proper feature selection strategy on the best $m$ channels, further improvement of the the classification accuracy can be expected. Moreover, there exists other features,~\emph{e.g.}, Fourier features, that can be used for EEG classification. Although the Fourier features does not fit the definition of Dirichlet distribution naturally, we can apply proper normalization strategy to make the feature neutral. Since Fourier features are more intuitive, classification accuracy improvement with normalized neutral Fourier feature can also be expected.

%%%%%%%%%%%%%%%%%%%%%%%%%%%%%%%%%%%%%%%%%%

\section{Acknowledgements}

The authors would like to thank the reviewers for their fruitful suggestions. Also, the authors would like to thank Dr. Jing-Hao Xue for his kind discussions and suggestions.

This work was partly supported by the National Natural Science Foundation of China (NSFC) under grant No.~$61402047$ and No.~$61273217$, the Scientific Research Foundation for Returned Scholars, Ministry of Education of China, Chinese $111$ program of Advanced Intelligence and Network Service under grant No.~B$08004$, and EU FP$7$ IRSES MobileCloud Project (Grant No.~$612212$).

\bibliographystyle{IEEEtran}
%\bibliography{JNL16}

\end{document}